\documentclass[letterpaper, 10 pt, journal, twoside]{ieeetran} 
\usepackage{hhline}
\usepackage{graphicx}
\usepackage{caption}
\usepackage{subcaption}
\usepackage{multirow}
\usepackage{soul,color}
\usepackage{amsfonts}
\usepackage{float}
\usepackage{amsmath}
\usepackage{algorithm}
\usepackage{algorithmic}
\usepackage{boldline} 

\usepackage{kotex}
\usepackage{kotex-logo}

\IEEEoverridecommandlockouts                              

\title{Gaze-based dual resolution deep imitation learning for high-precision dexterous robot manipulation}
\author{Heecheol Kim$^{1}$, Yoshiyuki Ohmura$^{1}$, and Yasuo Kuniyoshi$^{1}$%
\thanks{Manuscript received: October, 15, 2020; Revised January, 6, 2021; Accepted January, 25, 2021.}
\thanks{This paper was recommended for publication by Editor Dana Kulic upon evaluation of the Associate Editor and Reviewers' comments. 
This paper is based partly on results obtained under a Grant-in-Aid for Scientific Research (A) JP18H04108 and partly on results obtained from a project commissioned by the New Energy and Industrial Technology Development Organization (NEDO).} 
\thanks{$^{1}$Heecheol Kim, Yoshiyuki Ohmura, and Yasuo Kuniyoshi are with Laboratory for Intelligent Systems and Informatics, Graduate School of Information Science and Technology, The University of Tokyo, 7-3-1 Hongo, Bunkyo-ku, Tokyo, Japan
        {\tt\footnotesize \{h-kim, ohmura, kuniyosh\}@isi.imi.i.u-tokyo.ac.jp}}%
\thanks{Digital Object Identifier (DOI): see top of this page.}
}
\begin{document}
\maketitle

\begin{abstract}
A high-precision manipulation task, such as needle threading, is challenging.
Physiological studies have proposed connecting low-resolution peripheral vision
and fast movement to transport the hand into the vicinity of an object, and 
using high-resolution foveated vision to achieve the accurate homing of the hand to the object.
The results of this study demonstrate that a deep imitation learning based method, inspired by the gaze-based dual resolution visuomotor control system in humans, can solve the 
needle threading task. First, we recorded the gaze movements of a human operator who was teleoperating a robot.
Then, we used only a high-resolution image around the gaze to precisely control the thread position when it was close to the target. 
We used a low-resolution peripheral image to reach the vicinity of the target.
The experimental results obtained in this study demonstrate that the proposed method enables precise manipulation tasks using a general-purpose robot manipulator and improves computational efficiency.

\end{abstract}
\providecommand{\keywords}[1]{\textbf{\textit{Index terms---}} #1}



\section{Introduction}
\IEEEPARstart{D}{eep} imitation learning (\cite{zhang2018deep,yang2016repeatable}) is used to train deep neural networks on demonstration data, and has good potential for application to robots used in daily life because it does not require a hard-coded robot control rule.
This study demonstrates that deep imitation learning can be applied to high-precision visuomotor robot manipulation tasks, such as needle threading. 
This task is difficult both for humans and robots because (1) the clearance is small, (2) the thread is deformable, and (3) the posture of the needle varies in each picking trial. Therefore, this task requires a complex control policy with high precision.
If it can be demonstrated that the general-purpose robot manipulator can learn a complex policy, the scope of future robots used in daily life will expand.

\begin{figure}
  \centering
  \vspace{0.0in}
  \begin{subfigure}[t]{0.25\textwidth}
    \captionsetup{width=.8\linewidth}
    \captionsetup{justification=centering}
    \includegraphics[width=0.98\linewidth]{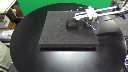}
    \caption{Low-resolution peripheral vision}
    \label{fig:small_left_gaze}
  \end{subfigure}
  \begin{subfigure}[t]{0.25\textwidth}
    \captionsetup{width=.8\linewidth}
    \captionsetup{justification=centering}
    \includegraphics[width=0.98\linewidth]{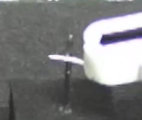}
    \caption{High-resolution foveated vision}
    \label{fig:fovea_left}
  \end{subfigure}
  \captionsetup{justification=centering}
  \caption{
  The proposed method can efficiently calculate a precise policy with both global features from peripheral vision (\ref{fig:small_left_gaze}) and detailed visual information for the needle and thread from foveated vision (\ref{fig:fovea_left}).
  }
  \label{fig:visual_img}
\end{figure}
Humans use two control systems with different visual resolutions for manipulation tasks \cite{paillard1996fast}. The anatomical study of the human eye indicates that the retina is divided into the fovea and peripheral vision. The fovea is located at the center of the retina with a cone photoreceptor-dominated field, which is denser and relatively thicker than other parts of the retina \cite{Kirkwood1995simple}, and thus provides high-resolution visual information. When a human moves its hand to a target, the eye gaze is naturally foveated at the target \cite{hayhoe2005eye,de2018visuomotor,sailer2005eye,safstrom2014gaze}, whereas the peripheral field contains a relatively sparse placement of photoreceptors, mostly dominated by rods \cite{Kirkwood1995simple}. Central vision and peripheral vision are functionally separated when the hand reaches out to the target under visual guidance \cite{paillard1996fast}. The human first moves its hand into the vicinity of the target with fast feedback loop control using peripheral visual information, and accomplishes the accurate homing of the target through a slow feedback loop with central vision at the end of the manipulation trajectory \cite{paillard1996fast,sarlegna2004online}.
However, most existing deep learning based visuomotor control methods for robot manipulation
learn an action policy with the entire visual input \cite{zhang2018deep,yang2016repeatable,finn2016deep,levine2016end}.
A recent study \cite{kim2020using} used human gaze during demonstrations to learn the action policy. However, the visuomotor control system was not separated into foveated vision and peripheral vision.

This study proposes a highly precise object manipulation method inspired by the separated visuomotor control system of humans.
The foveated vision ($10^\circ$) around the gaze position measured by the eye-tracker retains high spatial resolution ($142 \times 120$ from the $1280 \times 720$ entire image), and the entire image is resized as a low-resolution image ($128 \times 72$) to form the peripheral vision (Fig. \ref{fig:visual_img}). The deep neural network model infers fast-action only with a peripheral image to reach the target, and slow-action only with a foveated image to precisely grasp the thread or insert it into the eye of the needle. This separated visuomotor control architecture has two benefits compared with previous deep imitation learning methods. First, using the foveated vision with an explicit gaze mechanism improves generalization because (1) this mechanism extracts important visual features to focus on from the entire scene, and (2) fewer neural network parameters are required. Second, higher computational efficiency can be achieved because only the foveated vision retains high-resolution.

\section{Related work}
One representative task of small clearance manipulation is the peg-in-hole task. Because this task has small clearance, it is important to establish a control model of the peg under friction with the hole \cite{tang2016autonomous} or achieve the accurate alignment of the peg and hole \cite{inoue1974force,kim1999active,sharma2013intelligent,huang2013fast}. Force/torque feedback is mainly used to estimate the state of the peg/hole or the friction between the peg and the hole \cite{tang2016autonomous,majors1995neural,inoue2017deep}. \cite{huang2013fast} used visual feedback to align the peg and hole. 
By contrast, needle threading cannot use force feedback because the robot cannot sense friction between the needle and the flexible thread. Therefore, unlike the peg-in-hole which handles rigid objects, the robot must adjust its policy with thread deformation using visual feedback. Because it is difficult to establish a model with high-precision deformation prediction, the model-based approach cannot be applied to the needle threading task.

Related studies, such as \cite{inaba1985hand}, \cite{silverio2020laser} and \cite{Huang2015Robotic}, considered needle threading tasks. \cite{inaba1985hand} achieved a rope-into-ring task. An imaginary guide was set inside the ring to insert the rope tip into the ring. For the needle threading task considered in this study, it is difficult to calculate the imaginary guide tube because of the small clearance and insufficient visual resolution, even with $1280 \times 720$ images. Furthermore, in our setup, the robot must appropriately control the thread, even when the thread is deformed by collision with the needle. \cite{silverio2020laser} achieved high-precision assembly behaviors with laser-based sensing and validated with needle threading and  USB insertion tasks. A dual-arm system where one sensing arm mounts the laser scanner was used to compensate for errors. However, their approach requires a high-resolution depth sensing device mounted on a robot arm, which may not be suitable for a general-purpose robot system, and assumes that the thread is rigid.
\cite{Huang2015Robotic} achieved needle threading using high-speed visual feedback. Their idea was that a thread rotating with high-speed can be approximated as a rigid object, and thus the control models can be simplified. However, this study used a two-degree-of-freedom single purpose mechanism and did not automate the thread mounting. To the best of our knowledge, a series of pick-and-threading tasks has not been learned by a general-purpose robot manipulator to date.

Active vision is a research field about actively control the camera coordinate with respect to given sensory input to focus visual attention to a relevant target \cite{aloimonos1988active,bajcsy1988active,ballard1991animate}.
\cite{soong1991inverse,crowley1992gaze} achieved active stereo vision on a camera head system. \cite{sandini1980anthropomorphic,kuniyoshi1995active,kuniyoshi1995foveated} designed camera systems in which high-resolution central vision and low-resolution peripheral vision in hardware, and \cite{kuniyoshi1995active,kuniyoshi1995foveated} further implemented the proposed camera lens on an active stereo vision system.
\cite{whitehead1991learning,zhang2018agil} extended attention to sensory-motor control. \cite{whitehead1991learning} proposed a reinforcement learning system that can learn to focus on necessary sensory input. \cite{zhang2018agil} used foveated images for imitation learning on a two-dimensional video game.
However, our research objective is to demonstrate that visual attention acquired from the human gaze can improve performance for real-world robot manipulation.

\begin{figure}
  \centering
  \vspace{0.1in}
  \begin{subfigure}[t]{0.25\textwidth}
    \captionsetup{width=.8\linewidth}
    \captionsetup{justification=centering}
    \includegraphics[width=0.98\linewidth]{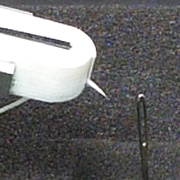}
    \caption{Short grasp}
    \label{fig:d1}
  \end{subfigure}%
  \begin{subfigure}[t]{0.25\textwidth}
    \captionsetup{width=.8\linewidth}
    \captionsetup{justification=centering}
    \includegraphics[width=0.98\linewidth]{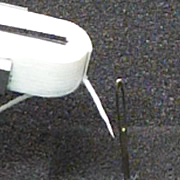}
    \caption{Long grasp}
    \label{fig:d2}
  \end{subfigure}%
  \captionsetup{justification=centering}
  \caption{
  Difference between thread grasps. The proposed method can adjust its policy with respect to the posture of the grasped thread.
  }
  \label{fig:diff}
\end{figure}

\section{Method}

\begin{figure}
  \centering
  \includegraphics[width=0.98\linewidth]{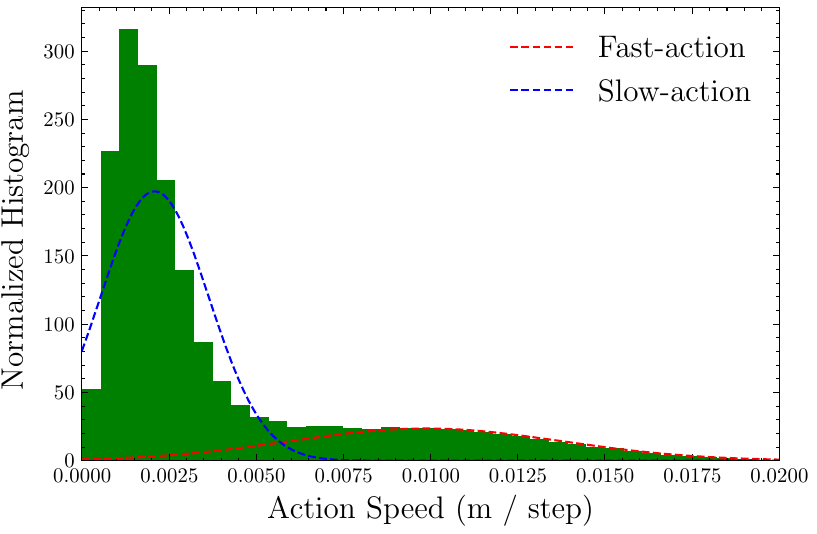}
  \captionsetup{justification=centering}
  \caption{
  Histogram of action speed and fitted GMM of needle threading. The intersection point of the two Gaussian distributions is defined as the threshold between slow-action and fast-action.
  }

  \label{fig:hist}
\end{figure}

\begin{figure*}
  \centering
  \vspace{0.02in}
  \includegraphics[width=.9\linewidth]{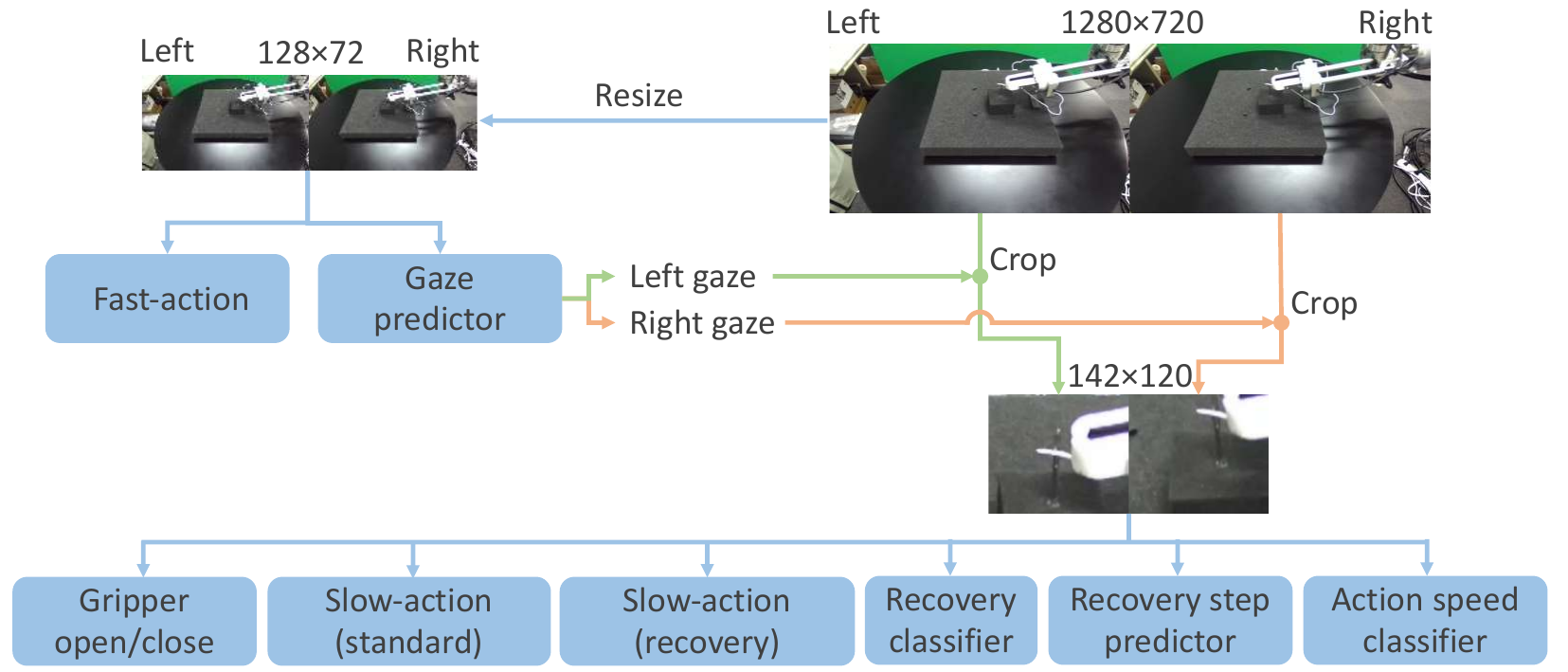}
  \captionsetup{width=.9\linewidth}
  \captionsetup{justification=centering}
  \caption{Proposed architecture.}
  \label{fig:network_architecture}
\end{figure*}

\begin{algorithm}[tb]
\caption{Proposed algorithm}
\label{alg:algo}
\textbf{Parameter}: Fast-action $\pi_{fast}$, slow-action (standard) $\pi_{slow}$, slow-action (recovery) $\pi_{rec}$, gripper open/close $\pi_{gp}$, recovery classifier $\theta$, gaze predictor $\rho$,  recovery step predictor $\psi$, action speed classifier $\phi$ \\
\begin{algorithmic}[1] 
\STATE $step \gets 0$
\STATE $count \gets 0$
\STATE $succeed \gets $ false
\WHILE{$step < 500$ \AND $\neg$ $succeed$}
    \STATE $o_l$, $o_r$ $\gets$ $1280 \times 720$ left/right camera image
    \STATE $p_l, p_r \gets resize(o_l, (128, 72)), resize(o_r, (128, 72))$ \COMMENT{resize to left/right peripheral vision $p_l$, $p_r$}
    \STATE $p \gets concat(p_l, p_r)$ \COMMENT{concatenate peripheral vision}
    \STATE $g_l, g_r \gets \rho(p)$ \COMMENT{predict left/right gaze position $g_l, g_r$}
    \STATE $c_l, c_r \gets crop(o_l, g_l), crop(o_r, g_r)$ \COMMENT{crop left/right foveated vision}
    \STATE $c \gets concat(c_l, c_r)$ \COMMENT{concatenate foveated vision}
    \STATE $gp \gets \pi_{gp}(c)$ \COMMENT{predict gripper command $gp$}
    \IF{$\theta(c)$ is true \AND $count = 0$}
        \STATE $count \gets \psi(c)$
    \ENDIF
    \IF{$\phi(c)$ is fast-action}
        \STATE $action \gets \pi_{fast}(p)$ \COMMENT{predict action from peripheral vision}
    \ELSE
        \IF{$count > 0$}
            \STATE $action \gets \pi_{rec}(c)$\COMMENT{predict recovery action from foveated vision}
            \STATE $count \gets count - 1$
        \ELSE
            \STATE $action \gets \pi_{slow}(c)$ \COMMENT{predict action from foveated vision}
        \ENDIF
    \ENDIF
    \STATE Execute $(action,gp)$ on robot 
    \STATE $step \gets step + 1$
    \STATE Manually decide $succeed$
\ENDWHILE
\end{algorithmic}
\end{algorithm}

\subsection{High-resolution image processing with human gaze}\label{subsection:gaze_processing}
In our setup, a human operator teleoperates a UR5 robot (Universal Robots) based on visual input from a head-mounted display (HMD), which reflects images from the ZED Mini stereo camera \cite{ZED-MINI} on the robot, while the eye tracker mounted on the HMD measures the operator’s eye gaze.   

The high definition (HD) stereo image ($1280 \times 720$) from the camera is processed into a foveated image and peripheral image. The peripheral image is the entire stereo image reduced to $128 \times 72$. The foveated image is a stereo image cropped to $142 \times 120$ around both the left and right gaze from the raw HD image, and corresponds to the central retina ($10^\circ$) of the human eye \cite{paillard1996fast} (see Appendix \ref{appendix:gazed} for details).
The derived foveated and peripheral images are used as input for the visuomotor policy calculation introduced in \ref{subsection:model_architecture}.
Each image and subsequent robot joint angles are collected at 10Hz. Therefore, each step is defined as 0.1 seconds throughout this paper.

\subsection{Action separation by speed of action}\label{subsection:action_scale}
The target-reaching movement of humans is divided into the fast reaching movement calculated from peripheral vision and accurate homing with a slow feedback loop calculated from central vision \cite{paillard1996fast,sarlegna2004online}. In our method, on the basis of this result obtained from physiological studies, the separated foveated and peripheral images are used to infer slow-action and fast-action, respectively.
As rotation is not significant in our needle threading and bolt picking tasks, the action speed is defined by the Euclidean norm of the positional difference (Eq. \ref{eq:speed}), whereas the action is defined as the difference in the end-effector position and orientation between the next step and current step during the teleoperation: 
\begin{equation}\label{eq:speed}
    s_t = \sqrt{(x_{t+1} -x_t)^2 + (y_{t+1} - y_t)^2 + (z_{t+1} - z_t)^2,}
\end{equation}
where $s_t$ represents the speed of action at timestep $t$. The action is separated into fast-action and slow-action according to the speed of the action using a threshold.
To determine the threshold that separates fast-action and slow-action, we assume that fast-action and slow-action are sampled from mutually independent probability distributions in terms of the action speed. 
A Gaussian mixture model (GMM) with two Gaussian distributions corresponding to the slow-action and fast-action, respectively, is fitted from the $[x, y, z]$ speed of actions (Fig. {\ref{fig:hist}}). Then, the intersection point of the two Gaussian distributions is determined as the threshold.

After learning, the action has to be automatically selected from either slow-action or fast-action. To achieve this, a binary classifier, that is, the \textit{action speed classifier}, is trained using the action speed labels.
The foveated vision is used as the input to the \textit{action speed classifier} because we observed that the foveated vision had sufficient information for classification.


\subsection{Recovery action}\label{subsection:recovery}

\begin{figure}
  \centering
  \vspace{0.0in}
  \begin{subfigure}[t]{0.125\textwidth}
    \captionsetup{width=.8\linewidth}
    \captionsetup{justification=centering}
    \includegraphics[width=0.98\linewidth]{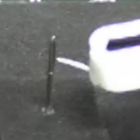}
    \caption{Approach}
    \label{fig:43}
  \end{subfigure}%
  \begin{subfigure}[t]{0.125\textwidth}
    \captionsetup{width=.8\linewidth}
    \captionsetup{justification=centering}
    \includegraphics[width=0.98\linewidth]{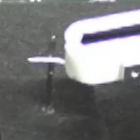}
    \caption{Failure}
    \label{fig:50}
  \end{subfigure}%
    \begin{subfigure}[t]{0.125\textwidth}
    \captionsetup{width=.8\linewidth}
    \captionsetup{justification=centering}
    \includegraphics[width=0.98\linewidth]{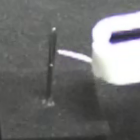}
    \caption{Recovery}
    \label{fig:59}
  \end{subfigure}%
  \begin{subfigure}[t]{0.125\textwidth}
    \captionsetup{width=.8\linewidth}
    \captionsetup{justification=centering}
    \includegraphics[width=0.98\linewidth]{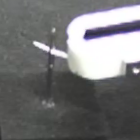}
    \caption{Success}
    \label{fig:69}
  \end{subfigure}%
  \captionsetup{justification=centering}
  \caption{
  Example of task failure caused by the human operator. The operator failed to thread the needle (\ref{fig:50}), recovered from the failure (\ref{fig:59}), retried, and finally succeeded (\ref{fig:69}).
  }
  \label{fig:failure_img}
\end{figure}

Owing to the small clearance of the needle threading task, even human demonstrators fail frequently. In the training data, the failure rate on each trial was 36.4\%.
In these failures, humans recover from the failure and try again until the threading is finally successful (Fig. \ref{fig:failure_img}). Hence, we assume that the robot should be able to recover from failure and achieve better performance in subsequent attempts.

In this study, we adopt neural networks that recognize the failure state and switch to the recovery action (Fig. \ref{fig:network_architecture}). The failure states and recovery actions are annotated based on demonstration data, respectively. The failure states are defined as threads passing by the needle without being inserted into the eye of the needle, whereas the recovery actions are defined as retreating actions until the next threading. Failure actions are removed from the training data.

The annotated recovery actions are separated from other threading actions in the training data (standard action) and are trained in a separate neural network that only infers the recovery action. Two additional neural networks are used to recognize the failure state. From the current foveated image, the \textit{recovery classifier} predicts whether the current state is a failure, and the \textit{recovery step predictor} predicts how many steps are required to complete the recovery. If the current state is a failure, the robot executes the inferred recovery action for the predicted number of steps. During the recovery steps, the robot continues the recovery action regardless of the change of the \textit{recovery classifier}’s result.

\subsection{Model architecture}\label{subsection:model_architecture}

\begin{figure}
  \centering
  \vspace{0.0in}
  \begin{subfigure}[t]{0.166\textwidth}
    \captionsetup{width=.8\linewidth}
    \captionsetup{justification=centering}
    \includegraphics[width=0.98\linewidth]{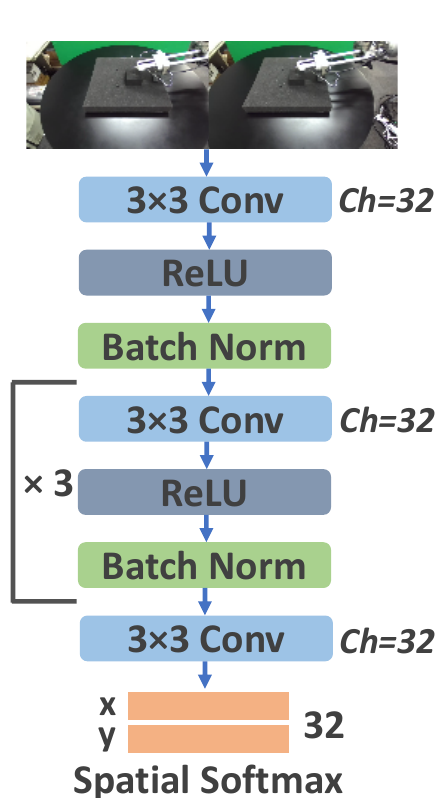}
    \caption{Peripheral vision processing architecture}
    \label{fig:global_network}
  \end{subfigure}%
  \begin{subfigure}[t]{0.166\textwidth}
    \captionsetup{width=.8\linewidth}
    \captionsetup{justification=centering}
    \includegraphics[width=0.98\linewidth]{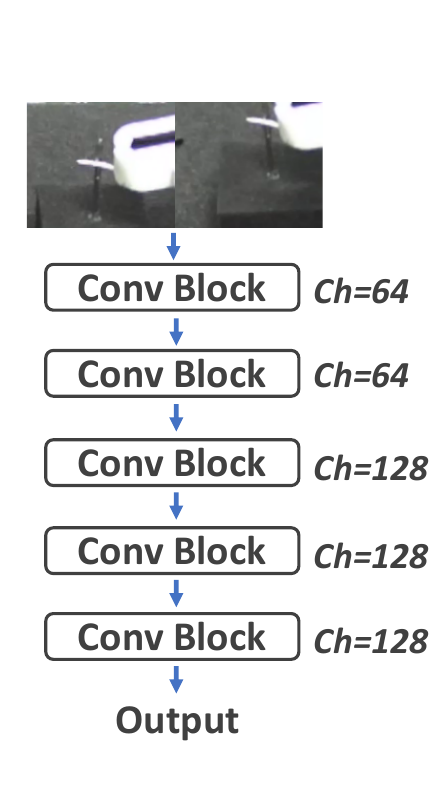}
    \caption{Foveated vision processing architecture}
    \label{fig:local_network}
  \end{subfigure}%
  \begin{subfigure}[t]{0.166\textwidth}
    \captionsetup{width=.8\linewidth}
    \captionsetup{justification=centering}
    \includegraphics[width=0.98\linewidth]{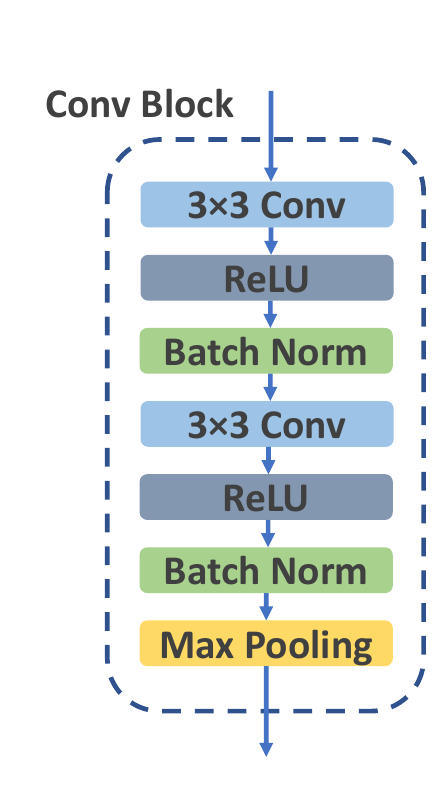}
    \caption{Convolution block}
    \label{fig:conv_block}
  \end{subfigure}%
  \captionsetup{justification=centering}
  \caption{Neural network architectures.}
  \label{fig:networks}

\end{figure}

It is important to recognize the three-dimensional state of the thread for the needle threading task. Therefore, both the left and right image of the stereo camera are used as input.

The entire architecture is shown in Fig. \ref{fig:network_architecture}. First, the raw $1280 \times 720$ left and right RGB images are concatenated into six channels and resized to a $128 \times 72$ peripheral image. This peripheral image is used to infer the fast-action and left/right gaze coordinate. The left/right foveated images are cropped from left/right $1280 \times 720$ raw images using the inferred left/right gaze coordinate, respectively. The left/right foveated images are concatenated into six channels. 
The foveated image is used to infer (1) the gripper opening/closing, (2) standard slow-action, (3) recovery slow-action, (4) recovery classification, (5) recovery step prediction, and (6) action speed classification (slow/fast) (Fig. \ref{fig:network_architecture}). The \textit{action speed classifier} determines whether to use the slow-action or fast-action in the current state. 
The \textit{recovery classifier} predicts whether the robot has failed to thread the needle. If the classifier decides to use the slow-action and recovery is required in the current state, the recovery action is selected and executed for $n$ steps, where $n$ is the number of steps predicted by the \textit{recovery step predictor}. If recovery is not required, a standard (fast or slow) action is executed based on the decision of the \textit{action speed classifier} (Algorithm \ref{alg:algo}).

Similar to previous research \cite{kim2020using}, gaze coordinate is inferred by the mixture density network (MDN) \cite{bishop1994mixture}. The MDN computes the parameters of the GMM to estimate the probability distribution of the target conditioned on the input data. In this study, the MDN architecture inputs a concatenated left/right peripheral image at the current time step to output $\mu$, $\sigma$, and $\rho$, which represent the mean, standard deviation, and correlation of the two-dimensional gaze position probability, respectively. 

The coordinate of the end-effector and the target is important for fast-action control using the peripheral image because the fast-action control moves the end-effector into the vicinity of the target.
To extract the coordinate information as a feature, spatial softmax \cite{zhang2018deep,finn2016deep}, which represents a feature as a two-dimensional coordinate (Fig. \ref{fig:global_network}), is used. The gaze predictor also uses the network architecture with the spatial softmax (Fig. \ref{fig:global_network}) because the target location is important in gaze coordinate prediction.

By contrast, the recognition of the three-dimensional state of the thread and the needle is important for the processing of the foveated image.
Spatial softmax is inadequate for inferring such information because the extracted feature is a two-dimensional coordinate. Therefore, the foveated image is processed using a series of convolutional neural networks and max-pooling layers with a stride of two (Fig. \ref{fig:local_network}, \ref{fig:conv_block}).

The extracted features are processed with a series of fully-connected (FC) layers. Batch normalization \cite{ioffe2015batch} and the ReLU activation function \cite{glorot2011deep} are used between the FC layers. There are 200 nodes for each FC layer. Four neural network models that infer actions (standard slow-action, recovery slow-action, fast-action, and gripper opening/closing) use three FC layers followed by Fig. \ref{fig:networks}, whereas the other neural network models use two FC layers.

\section{Experiments}

\begin{figure}
  \centering
  \vspace{0.0in}
  \begin{subfigure}[t]{0.25\textwidth}
    \captionsetup{width=.8\linewidth}
    \captionsetup{justification=centering}
    \includegraphics[width=0.98\linewidth]{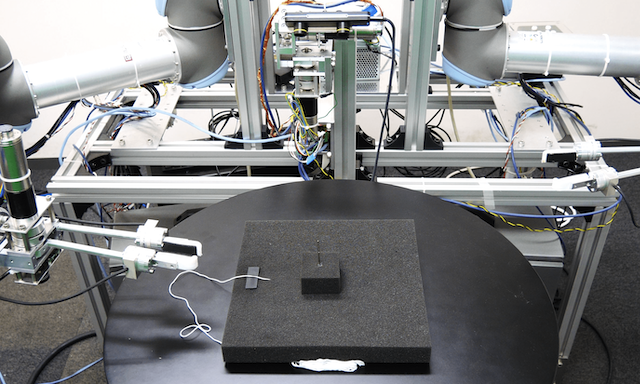}
    \caption{Needle threading}
    \label{fig:needle_threading}
  \end{subfigure}%
  \begin{subfigure}[t]{0.25\textwidth}
    \captionsetup{width=.8\linewidth}
    \captionsetup{justification=centering}
    \includegraphics[width=0.98\linewidth]{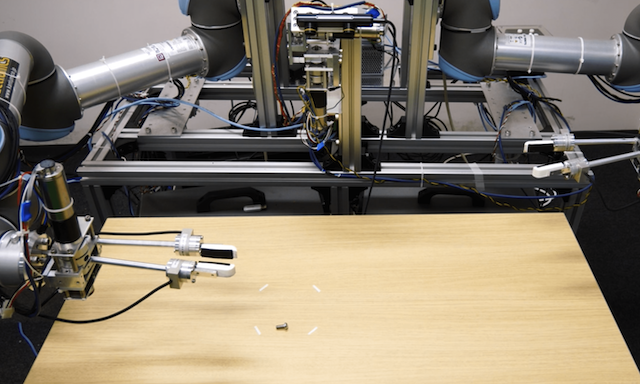}
    \caption{Bolt picking}
    \label{fig:bolt_picking}
  \end{subfigure}%
  \captionsetup{justification=centering}
  \caption{Task setup.}
  \label{fig:setup}
\end{figure}

\begin{figure*}
\centering 
\vspace{0.1in}
\includegraphics[width=1.0\linewidth]{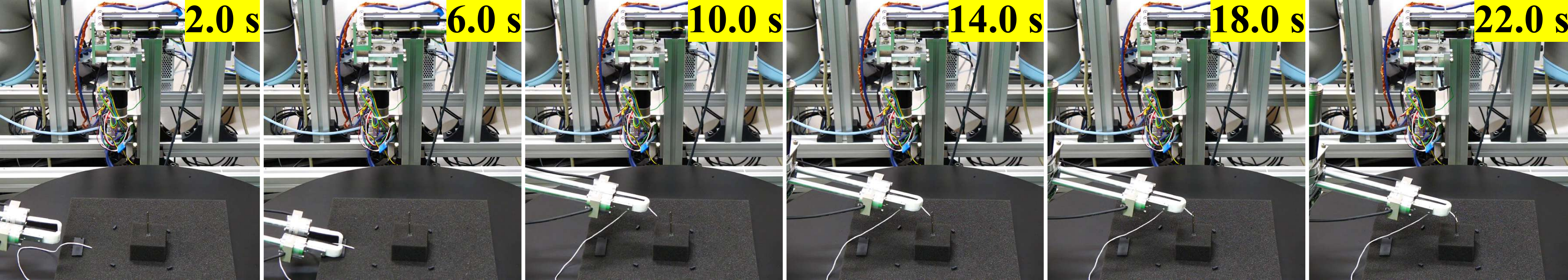}\par\medskip
\includegraphics[width=1.0\linewidth]{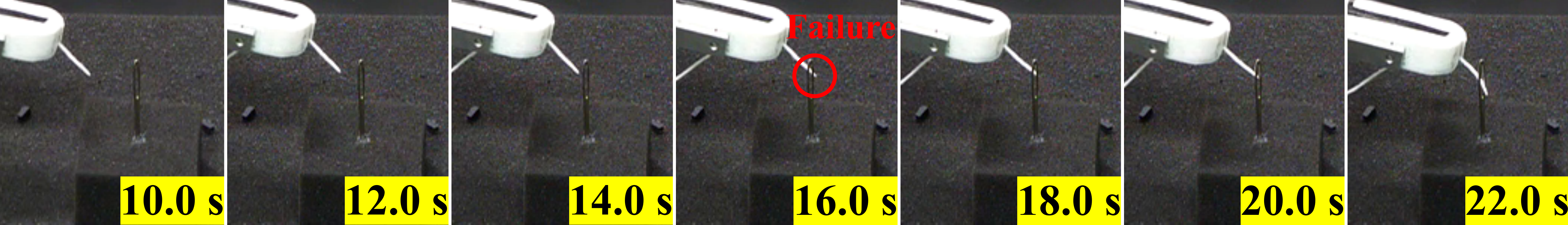}
\caption{
Example of a successful trial. The robot was able to recognize the failure (16.0 s), recover from it (18.0 s), and finally succeed in threading the needle.
}
\label{fig:trial0}
\end{figure*}

\begin{figure*}
    \centering
    \includegraphics[width=1.0\linewidth]{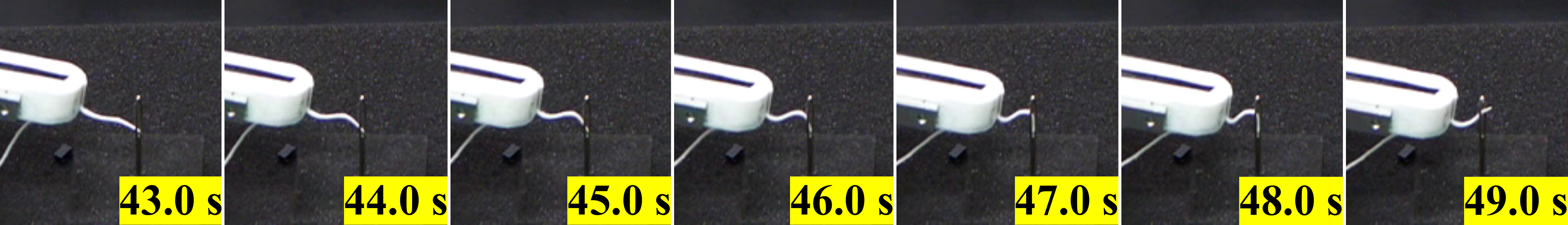}
    \caption{
    Another successful trial. The proposed method was able to perform manipulation, even when the thread was deformed.
    }
    \label{fig:trial2}
\end{figure*}

\subsection{Task setup}

In the needle threading task (Fig. \ref{fig:needle_threading}), the robot manipulator had to (1) pick up the thread and (2) insert it into the eye of the needle that was randomly placed on a $9 cm \times 9 cm$ area on the table. 
In this setup, the trials varied in the location of the needle and the grasping position of the thread (Fig. \ref{fig:diff}). The neural network had to adjust its policy with such variances.

To demonstrate that the proposed method is generally applicable to precise tasks, a bolt-picking task was also conducted (Fig. \ref{fig:bolt_picking}). In this task, the robot had to pick up a $M6 \times 14$ bolt. Hence, the robot had to learn the precise location to grasp the bolt while avoiding colliding with the table.

\begin{table*}[]
\centering
\begin{tabular}{llll}
\hlineB{2}
Experiment                                                       & Method                                                              & Pick (16 trials) (\%)           & Thread (16 trials) (\%)         \\ \hline \hline
(a) Assessment of the proposed method                                & Proposed method                                                     & 93.75                     & 81.25                     \\ \hline
\multirow{2}{*}{(b) Assessment of foveated vision}               & Without gaze (image size: $128 \times 72$)                          & 93.75                     & 12.50$^*$  \\
                                                                 & Without gaze (image size: $1280 \times 720$)                        & 0.00$^*$  & 0.00$^*$  \\ \hline
\multirow{2}{*}{(c) Assessment of action separation}             & Without action separation                                           & 43.75$^*$ & 0.00$^*$  \\
                                                                 & Action speed threshold = 0.00208                                    & 25.00$^*$ & 12.50$^*$ \\
                                                                 & Action speed threshold = 0.00364                                    & 93.75                     & 68.75                     \\
                                                                 & Action speed threshold = 0.00790                                    & 93.75                     & 50.00                     \\
                                                                 & Action speed threshold = 0.00985                                    & 87.50                     & 43.75$^*$ \\ \hline
\multirow{3}{*}{(d) Assessment of visuomotor control separation} & Fast-action with foveated vision only         & 0.00$^*$  & 0.00$^*$  \\
                                                                 & Fast-action with foveated + peripheral vision & 6.25$^*$  & 0.00$^*$  \\
                                                                 & Slow-action with foveated + peripheral vision & 100.00                    & 62.50                     \\ \hline
\multirow{2}{*}{(e) Assessment of recovery action}               & Standard \& recovery action in one network    & 93.75                     & 50.00                     \\
                                                                 & No \textit{recovery step predictor}                & 93.75                     & 31.25$^*$ \\ \hline
(f) Assessment of resolution               & Half resolution ($640 \times 360$)                                  & 100.00                    & 50.00                     \\ \hline
(g) Assessment of stereo vision                                  & Left image only                                                     & 68.75                     & 12.50$^*$ \\ \hlineB{2}
\end{tabular}
\caption{Needle threading result. The proposed method was significantly better than the task results marked using $^*$ (chi-square test, $P < 0.05$).}\label{tab:result}
\end{table*}

\begin{table}[]
\centering
\begin{tabular}{lll}
\hlineB{2}
    Method                   & MACs (G) & Learning time (hours) \\ \hline \hline
Proposed method                  & 7.21     & 7.92                  \\
Without gaze ($1280 \times 720$) & 58.3     & 220                   \\ \hlineB{2}
\end{tabular}
\caption{Efficient learning of neural networks using the gaze (needle threading).}
\label{tab:compute_time}
\end{table}

\begin{table}[]
\centering
\begin{tabular}{lll}
\hlineB{2}
Method                           & Pick (16 trials) (\%) \\ \hline \hline
(a) Proposed method                  & 68.75          \\
(b) Without gaze (image size: $1280 \times 720$)   & 18.75           \\
(c) Without action separation          & 0.00           \\ \hlineB{2}
\end{tabular}
\caption{Bolt picking result.}
\label{tab:bolt_exp}
\end{table}


\subsection{Assessment of the proposed method}\label{subsection:proposed_method}
The proposed method achieved a success rate of 81.25\% for needle threading using 203.8 minutes of the training data and 68.75\% in bolt picking using 39.56 minutes of the training data (Tables \ref{tab:result}a and \ref{tab:bolt_exp}a, respectively). Fig. \ref{fig:trial0} illustrates that the robot grasped the thread (6.0 s), recognized the failure (16.0 s), recovered from the failure state (18.0 s), and finally succeeded in threading the needle (22.0 s). In another trial (Fig. \ref{fig:trial2}), the robot was able to perform the appropriate manipulation, even when the thread deformed after colliding with the needle. Fig. \ref{table:mdn_accuracy} shows the gaze predictor accuracy.

\begin{table}[]
\centering
\begin{tabular}{lllll}
\hlineB{2}
Dataset       & \multicolumn{2}{c}{Left (\%)}              & \multicolumn{2}{c}{Right (\%)} \\
               & Horizontal & \multicolumn{1}{l|}{Vertical} & Horizontal      & Vertical     \\ \hline \hline
Training set      & 0.670      & \multicolumn{1}{l|}{1.69}     & 0.639           & 1.10         \\
Validation set & 0.620      & \multicolumn{1}{l|}{1.61}     & 0.672           & 1.04         \\ \hlineB{2}
\end{tabular}
\caption{Median of the gaze prediction error (horizontal: $error\_in\_pixels / width \times 100$, vertical: $error\_in\_pixels / height \times 100$).}
\label{table:mdn_accuracy}
\end{table}

\subsection{Assessment of foveated vision}\label{subsection:gaze_effect}
To evaluate how the gaze affects both computational cost and performance, the proposed method was compared with methods that do not use foveated vision. 
When the foveated images were replaced with the peripheral image ($128 \times 72$), the neural networks failed at needle threading (success rate of 12.5\%). Moreover, when all images were replaced with the raw image ($1280 \times 720$), the neural networks even failed at picking a thread (Table \ref{tab:result}b).
This network architecture also failed at bolt picking (Table. \ref{tab:bolt_exp}b). Therefore, selecting an image with an appropriate resolution in accordance with the target task contributes to the improvement of the generalization.

The computational cost of the proposed method and that of processing a raw $1280 \times 720$ image were compared (Table. \ref{tab:compute_time}). The proposed method required 8.10 times fewer multiply-accumulate operations (MACs) to calculate one step, and approximately 27.7 times less computational time to train the entire 30 epochs of all neural networks. A difference between the saved computational time and the saved MACs was caused by overheads such as memory copy.


\subsection{Assessment of the action separation}\label{subsection:action_scale_effect}

The success rate of needle threading dropped to 0\% when both the fast-action and slow-action were trained using only the peripheral image. Additionally, the threshold search results confirmed that the threshold that separates slow-action from fast-action should be the intersection point of the two Gaussian distributions (Tables \ref{tab:result}c and \ref{tab:bolt_exp}c). In addition to the intersection point of $0.005685 \sim \mu_1 + 2.315\sigma_1 \sim \mu_2 $-$ 1.069 \sigma_2$, four possible thresholds were tested: $[0.00208, 0.00364, 0.00790, 0.00985]$, which correspond to $[ \mu_1, \mu_1 + \sigma_1, \mu_2 $-$ \frac{\sigma_2}{2}, \mu_2]$, respectively. $\mu_1, \mu_2$, and $\sigma_1, \sigma_2$ are the mean and standard deviation of each Gaussian ($\mu_1 < \mu_2$), respectively. 

\subsection{Assessment of visuomotor control separation}\label{subsection:architecture_search}
The proposed method assigned peripheral vision to the fast-action and foveated vision to the slow-action. For other architectures, such as allocating only foveated vision to both the fast-action and slow-action or assigning both foveated and peripheral vision to the fast-action, the robot failed to pick up the thread. Moreover, allocating both foveated and peripheral vision to slow-action decreased the success rate (Table \ref{tab:result}d). 

\subsection{Assessment of the recovery action}\label{subsection:recovery_action_effect}
Training both standard and recovery actions using only one network resulted in inferior performance (Table \ref{tab:result}e). The reason for this is that the recovery action was not explicitly acquired, thus the robot could not recover after failure. Notably, the proposed method was able to recover from 12 out of 13 failures, whereas the neural network trained with both standard and recovery actions only recovered from one out of eight failures.
Additionally, the success rate dropped to 31.25\% without the \textit{recovery step predictor}. At this time, the robot repeatedly switched between the standard action and the recovery action.

\subsection{Assessment of the resolution}\label{subsection:high_resolution_effect}
When a half-resolution image ($640 \times 360$) was used, the success rate dropped to 50\% (Table \ref{tab:result}f), which demonstrates that high-resolution images are required to conduct needle threading tasks.

\subsection{Assessment of stereo vision}\label{subsection:setero_vision_effect}
To evaluate the necessity of stereo vision, the entire neural network architectures were re-trained with mono vision using images from the left camera only (therefore, the number of the input channels was three).
Without stereo vision, the success rate dramatically decreased (Table. \ref{tab:result}g).

\section{Discussion}
In this paper, we proposed a gaze-based dual resolution deep imitation learning method for robot manipulation. The proposed method separates the foveated and peripheral vision, which are used to control the slow-action and fast-action, respectively. The proposed method enabled a general-purpose robot manipulator to perform needle threading with high precision. This approach achieved appropriate control even when the thread was deformed. Additionally, the experimental results obtained for the bolt picking task partially demonstrated that the proposed method can be generalized and applied to other tasks that require high precision. As the proposed method enables the high-precision manipulation of the a deformable object only from human demonstration data, it is potentially applicable to tasks that require dexterous manipulation skills such as food processing, jewelry manufacturing, or clothing manipulation.

The results obtained in this study confirm that this separated visuomotor control improves both the task performance and computational efficiency compared with processing using high-resolution visual input. Moreover, the results of this research agree with the results obtained by previous studies on the role of central/peripheral vision in visuomotor manipulation \cite{paillard1996fast,de2018visuomotor,sarlegna2004online}. This suggests that the characteristics of human behavior must be considered in the application of imitation learning using human-generated data to robots, and that only using end-to-end fitting for the motor control output from input sensory data may be inadequate.

In this research, the foveated vision is defined as $10^\circ$ around the gaze position. The experimental results demonstrated that this setup can solve tasks that handle small objects. Whether the same setup can solve tasks with large objects still requires investigation. The area of the foveated vision can be related to the target objects. Therefore, the active adjustment of the foveated vision area in accordance with the target task may further increase the generality of the proposed method. Additionally, the stereo camera was fixed in this work. To fully use the advantages of active vision, active control of the camera head, in addition to gaze coordinate prediction in the image, is required.

\appendix
\subsection{Calculation of foveated vision from the camera image}\label{appendix:gazed}
The field of view of the ZED Mini stereo camera is $90^\circ (H) \times 60^\circ (V) \times 100^\circ$. Therefore, the central retina ($10^\circ$) corresponds to $142 \times 120$ of the image from the full image size of $1280 \times 720$.

\subsection{Training details}
The neural networks were trained on needle threading using 1,713 episodes of training data (203.8 minutes) and 148 episodes of validation data (17.01 minutes). 
For bolt picking, 775 training episodes (39.56 minutes) and 102 validation episodes (5.402 minutes) were used.
The demonstration was sampled at 10.0 Hz. Each neural network component was selected from the lowest validation loss over 30 epochs.
The learning rate of $1e-4$, rectified Adam (RAdam) optimizer \cite{liu2019variance}, and the same action loss as that in \cite{kim2020using} were used.

Each neural network component was trained using a Xeon CPU E5-2698 v4 and an NVIDIA Tesla V100 GPU with a batch size of 64. When training was conducted using a $1280 \times 720$ image, a batch size of 8 was used to avoid memory issues. An Intel CPU Core i7-8700K and one NVIDIA GeForce GTX 1080 Ti were used to control the UR5 robot manipulator. 

\subsection{Evaluation details}
The needle block was a $5cm \times 5cm$ sponge block with the needle installed in the middle, and was placed as similarly as possible to the previously recorded test positions.
To ensure a fair comparison of the test results, approximately $1 \sim 2 mm$ of the end of the thread was hardened using glue. Notably, this did not result in loss of deformability for the thread. In the additional thread experiment, wherein the thread was not hardened using glue but moistened using a small amount of water, 68.75\% of successful threading was recorded. 
A trial was assessed as a failure when it exceeded the maximum steps of 500.

\subsection{Classifier accuracy}
The classification accuracy of the \textit{recovery classifier} and \textit{action speed classier} are presented in Table. \ref{table:accuracy}.

\begin{table}[]
\centering
\begin{tabular}{lll}
\hlineB{2}
Classifier                & Training set (\%) & Validation set (\%) \\ \hline \hline
Recovery classifier     & 97.43   & 94.90        \\
Action speed classifier & 94.20   & 93.99        \\ \hlineB{2}
\end{tabular}
\caption{Classification accuracy.}
\label{table:accuracy}
\end{table}

\begin{figure}
  \centering
  \vspace{0.0in}
    \captionsetup{width=.8\linewidth}
    \captionsetup{justification=centering}
    \includegraphics[width=1.0\linewidth]{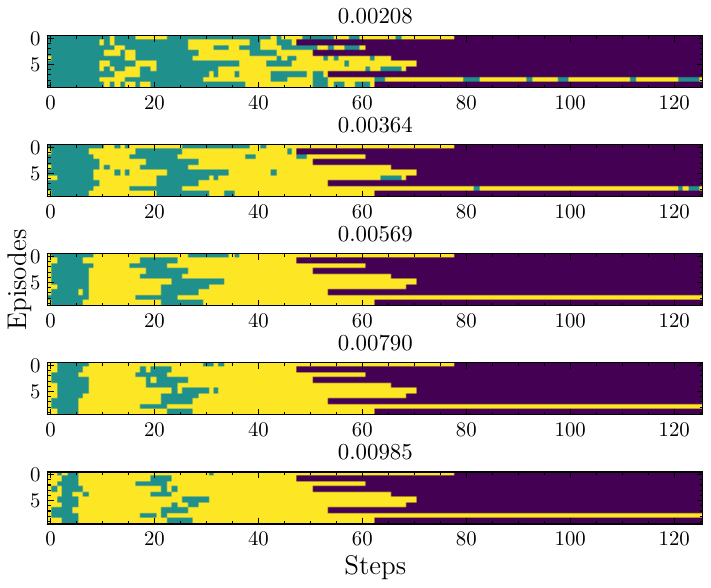}
    \caption{
    Fast/slow-action visualized on each threshold (horizontal: steps in an episode, vertical: episodes, \textcolor[rgb]{0.00392156862745098,0.22745098039215686,0.20784313725490197}{cyan}: fast-action, \textcolor[rgb]{1,0.8274509803921568,0}{yellow}: slow-action).
    }
    \label{fig:episode_trajs}
\end{figure}

\subsection{Assessment of the action separation threshold}
\begin{figure}
    \centering
    \captionsetup{width=.8\linewidth}
    \captionsetup{justification=centering}
    \includegraphics[width=1.0\linewidth]{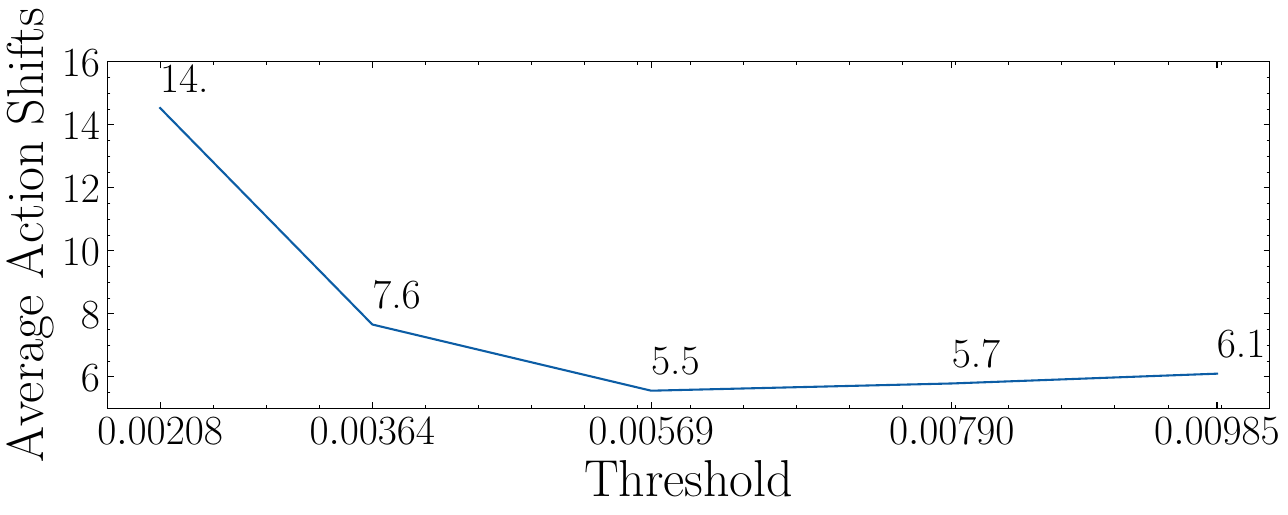}
    \caption{Average number action shifts (fast$\rightarrow$ slow \& slow $\rightarrow$ fast) on each threshold per episode. 
    }
    \label{fig:action_shift}
\end{figure}

Fig. \ref{fig:episode_trajs} visualizes fast-action (cyan) and slow-action (yellow) separated with thresholds used in \ref{subsection:action_scale_effect} on the validation set of the needle threading task. 
Threshold$=0.00569$, which is the intersection point of the two Gaussian distributions, showed the least number of action shifts (Fig. \ref{fig:action_shift}), which indicates that the proposed method shows the least interleaved sub-segmentation of slow and fast actions. 

\bibliographystyle{bibtex/bst/IEEEtran}
\bibliography{bibtex/bib/IEEEfull,bibtex/bib/mybibfile}

\end{document}